\documentclass[fleqn,10pt]{wlscirep}
\usepackage[utf8]{inputenc}
\usepackage[T1]{fontenc}
\usepackage{lineno}
\usepackage{multirow}
\usepackage{subcaption}
\usepackage{hyperref}
\hypersetup{
    colorlinks=true,
    citecolor=Green,
    linkcolor=NavyBlue,
    filecolor=OrangeRed,      
    urlcolor=OrangeRed,
    pdfpagemode=FullScreen,
}

\title{TartanAviation: Image, Speech, and ADS-B Trajectory Datasets for Terminal Airspace Operations}

\author[1]{Jay Patrikar}
\author[1]{Joao Dantas}
\author[1]{Brady Moon}
\author[1]{Milad Hamidi}
\author[1]{Sourish Ghosh}
\author[1]{Nikhil Keetha}
\author[1]{Ian Higgins}
\author[1]{Atharva Chandak}
\author[2]{Takashi Yoneyama}
\author[1]{Sebastian Scherer}

\affil[1]{Robotics Institute, Carnegie Mellon University}
\affil[2]{Aeronautics Institute of Technology, Sao Jose dos Campos, SP, Brazil}

\begin{abstract}

We introduce TartanAviation, an open-source multi-modal dataset focused on terminal-area airspace operations. TartanAviation provides a holistic view of the airport environment by concurrently collecting image, speech, and ADS-B trajectory data using setups installed inside airport boundaries. The datasets were collected at both towered and non-towered airfields across multiple months to capture diversity in aircraft operations, seasons, aircraft types, and weather conditions. In total, TartanAviation provides 3.1M images, 3374 hours of Air Traffic Control speech data, and 661 days of ADS-B trajectory data. The data was filtered, processed, and validated to create a curated dataset. In addition to the dataset, we also open-source the code-base used to collect and pre-process the dataset, further enhancing accessibility and usability. We believe this dataset has many potential use cases and would be particularly vital in allowing AI and machine learning technologies to be integrated into air traffic control systems and advance the adoption of autonomous aircraft in the airspace.

\textbf{Website:} \href{https://theairlab.org/tartanaviation/}{https://theairlab.org/tartanaviation/} $\mid$ \textbf{Code:} \href{https://github.com/castacks/TartanAviation.git}{https://github.com/castacks/TartanAviation}  

\end{abstract}

\begin{document}

\flushbottom
\maketitle

\thispagestyle{empty}


\section*{Background \& Summary}
\begin{figure}[ht]
    \centering
    \includegraphics[width=0.9\textwidth]{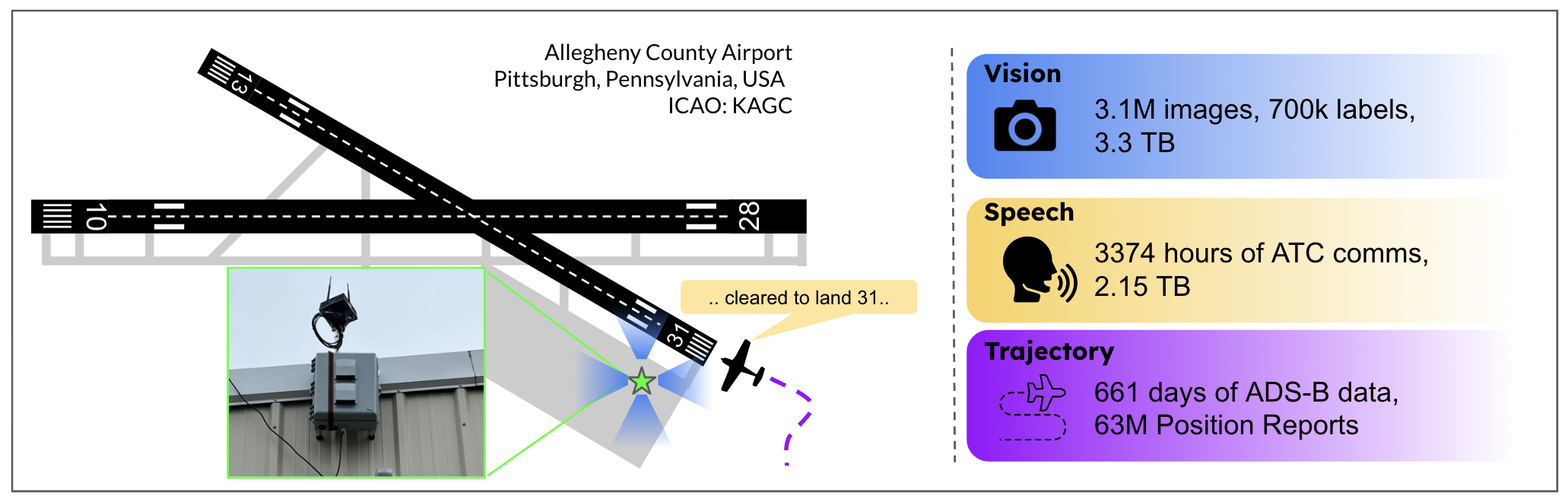}
    \caption{Our custom data collection setup installed at the Allegheny County Airport with its approximate location within the airport premises with respect to the runway geometry. The setup recorded images, audio, and aircraft trajectory data.}
    \label{fig:splash}
\end{figure}

In 2023, an average of 45,000 flights per day served over 2.9 million passengers in the US \cite{federal2023faa}, which is nearly at the saturation capacity of the current US air traffic control system \cite{mueller2017enabling}. The expected introduction of Advanced Aerial Mobility (AAM) operations within the National Airspace System heralds a further increase in flight operations, necessitating a need to increase the airspace capacity~\cite{iata2022air}. Compared to en-route airspace, terminal areas experience a higher air traffic density as nearly all aerial operations start or end within these areas. The effective future management of this high-risk airspace necessitates insights into various aspects of air traffic management. In addition to manned AAM operation, data is needed to achieve advances in fully or partially autonomous AAM agents. Curated datasets within this domain not only enable analytical research into understanding how the system is currently managed but also help accelerate the development of novel technologies for future operations. More generally, the aviation domain presents newer challenges to widely applied technologies like vision-based object detection, speech-to-text translation, and time-series analytics. Advancements in AI and machine learning can potentially revolutionize air traffic control systems, ensuring safer and more efficient coordination for the ever-increasing number of flights.


 In this work, we introduce TartanAviation, a multi-modal dataset collected at towered and non-towered terminal areas within the US. The TartanAviation dataset covers three primary concurrently collected modalities of data: trajectory positions for capturing the spatial and temporal information of aircraft movements, video flight sequences collected with static cameras installed within terminal areas, and audio communications to document the voice interactions between pilots and air traffic controllers. While prior datasets in the aviation domain have focused on specific modalities like speech \cite{lin2023towards,zuluaga2022atco2} or vision \cite{smyers2023avoidds,aotchallenge}, TartanAviation aims to provide a more holistic view of terminal airspace operations across various data modalities. Additionally, while previous datasets focus on large commercial airports, TartanAviation focuses on smaller regional airports within the Greater Pittsburgh area. Regional airports serve a multitude of different aircraft and mission profiles, providing a richer and more diverse data stream. We specifically focus on two airports: the towered Allegheny County Airport (ICAO:KAGC) \footnote{\href{http://www.airnav.com/airport/kagc}{http://www.airnav.com/airport/kagc}} and the non-towered Pittsburgh-Butler Regional Airport (ICAO:KBTP)\footnote{\href{https://www.airnav.com/airport/kbtp}{https://www.airnav.com/airport/kbtp}}.  

Human vision is the last line of defense against mid-air collisions, making it critical for aviation safety~\cite{weibel2004safety}. With recent advances in computer vision technologies, visual detect-and-avoid (DAA) systems have shown promising results in detecting airborne objects at greater distances \cite{sourish2022airtrack}. Prior datasets for DAA either lack diversity \cite{aotchallenge}, are computer-generated \cite{smyers2023avoidds}, or are relatively small \cite{farhadmanesh2022general,martin2022dataset}.
In this context, the TartanAviation vision dataset is a large-scale real-world dataset collected by placing static cameras within the terminal area. TartanAviation currently offers 3.1 million images covering challenging scenarios like snow, mist, rain, varying cloud cover, and diverse aircraft types. The data is augmented with the associated weather and air traffic trajectory data. With over 700k aircraft labels, these challenging real-world scenarios promise to enable more research in robust computer vision techniques for long-range object detection.

Trajectory data represents the time-series information of aircraft positions operating within the terminal airspace. This data is collected by recording Automatic Dependent Surveillance-Broadcasts (ADS-B) using receivers placed within the terminal area. Our prior work, TrajAir \cite{patrikar2021predicting}, provided 111 days of ADS-B data at KBTP. TartanAviation extends this tally to a total of 661 days of data at both KBTP and KAGC. Beyond the aviation domain, trajectory data can enable more research in time-series forecasting, social trajectory prediction, and anomaly detection.           

Radio communications enable pilots and Air Traffic Controllers (ATC) to share time-sensitive intent and instructions over dedicated frequencies. The information included in these communications helps pilots and ATC to build situational awareness, enabling the safe operation of aircraft on the ground and in the air. While most speech datasets are unimodal \cite{godfrey1994air}, recent multi-modal datasets with text and trajectory \cite{guo2023m2ats} have focused exclusively on commercial aircraft operations at large airports. TartanAviation provides first-of-its-kind speech data at relatively smaller airports, including both towered and untowered airfields. Along with the diversity in speech data, TartanAviation also provides concurrent trajectory data, enabling novel research in multi-modal speech-to-text translation and speech-to-intent predictions \cite{srinivasamurthy2017semi} conditioned on external context.

All of the data in TartanAviation was collected with administrative support and prior authorization at both the KAGC and KBTP airports. TartanAviation represents our first step at building a centralized comprehensive multi-modal repository for aircraft data within the terminal airspace.


\section*{Methods}

\begin{figure}
    \centering
    \includegraphics[width=0.9\textwidth]{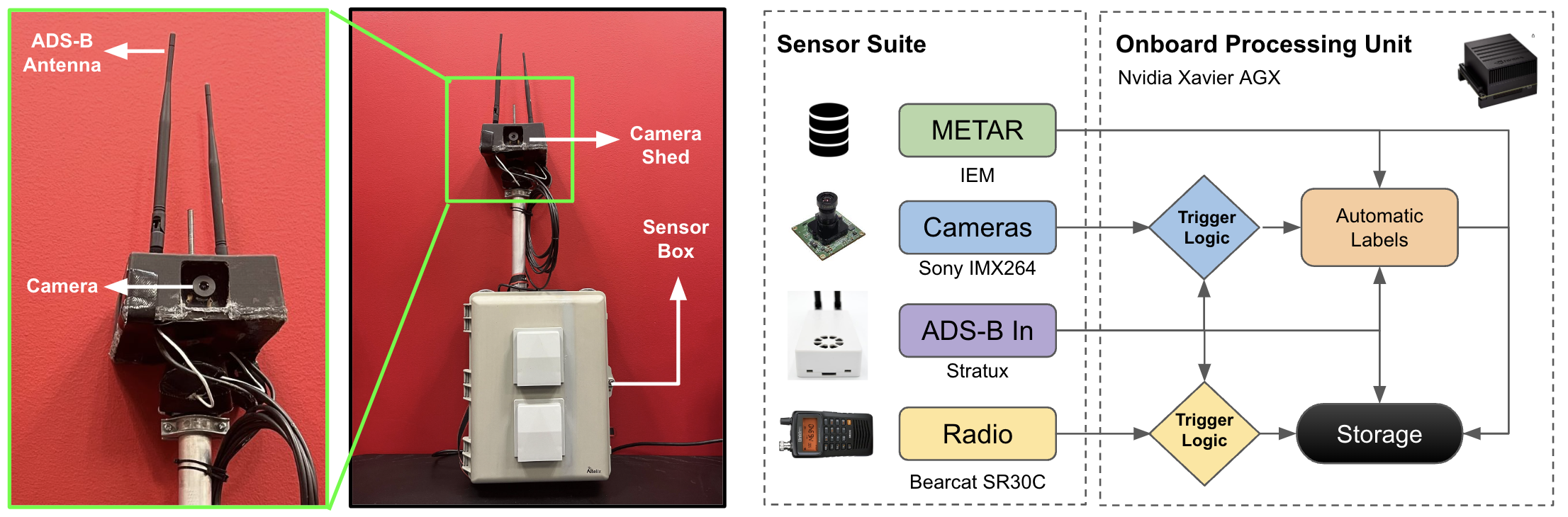}
    \caption{Our custom setup hardware with the camera and ADS-B antenna mounts \textit{(left)}. We also showcase the data collection pipeline with the associated sensor suite and automatic logic that triggers camera and speech recordings \textit{(right)}.}
    \label{fig:setup}
\end{figure}

This section briefly overviews the equipment and protocols used to collect the datasets. Figure \ref{fig:setup} shows the setup and an overview of the data collection pipeline.

\subsection*{Vision Data Acquisition}
The vision data was exclusively collected at the KAGC airport. The setup was mounted at 40.351422 latitude, -79.923939 longitude, as shown in Figure \ref{fig:splash}. The vision setup uses an array of Sony IMX 264 cameras connected to a NVIDIA Xavier AGX 32GB dev kit via a LI-JXAV-MIPI-ADPT-6CAM camera adaptor. The cameras collect data at $2048 \times 2448$ resolution at $24$ Frames Per Second (FPS). The camera recordings were triggered using ADS-B data by enforcing an 8 km fence around the setup. If an aircraft crosses the threshold, the camera automatically starts recording. The recording stops either at the end of a 250-sec timer or if the aircraft exits the 8 km fence. A new recording is activated if the recording times out before the aircraft exits. This process repeats till the aircraft leaves the geo-fence. To ensure seasonal diversity, data collection occurred in multiple phases from December 14, 2021, to February 23, 2023, from sunrise to sunset.

\subsubsection*{Post-Processing}

As shown in Table~\ref{tab:adsb}, along with the raw camera output, TartanAviation offers various useful meta-data, including ADS-B and bounding box labels.

For ADS-B, each recorded video sequence has a PKL file that contains the GPS coordinates of all flying aircraft in the area at the specific time of recording. 
The ADS-B data also provides the unique flight ID and N-number (if provided by the pilot to the ADS-B) per aircraft. 
The unique flight ID and the N-number of the aircraft are used to parse the FAA N-number registry inquiry website \footnote{\href{https://registry.faa.gov/aircraftinquiry/Search/NNumberInquiry}{https://registry.faa.gov/aircraftinquiry/Search/NNumberInquiry}} to get the aircraft type data.
Furthermore, we obtain the ADS-B data and interpolate it to obtain ground truth information matching the frequency of the camera recording, i.e., 24 Hz.

For the bonding box labels, each recorded video sequence has a zip file containing the label files for individual video frames. The labels are text files, where each row provides the bounding box information for a flying aircraft in the video. 
In particular, the label files include the bounding box coordinates for the aircraft in the image, a unique track ID for each aircraft, the range of the aircraft, the manufacturer, type, and model.

We leverage two main modules to generate the bounding box labels: (a) Auto-labeling and (b) Manual label verification/generation. 
The Auto-labeling module is performed in two stages.
The first stage uses our deep learning-based aircraft detection \& tracking system, AirTrack~\cite{sourish2022airtrack}, to obtain initial bounding box hypotheses.
Despite not having perfect recall, this stage makes labeling bounding boxes easier and reduces the amount of manual labor required downstream.
The second stage projects the 3D ADS-B data into the image frame using the calibrated extrinsics to provide an initial estimate of the aircraft position in the image.
This projection helps filter false positives from the detector and provides the initial airborne object location to the labeler.
Once the auto-labeling procedure is finished, we manually verify the image labels frame-by-frame and make any necessary corrections using our custom labeling software.

The system used in the Auto-labeling module, AirTrack~\cite{sourish2022airtrack}, consists of modules for ego-aircraft motion estimation, detection and tracking, and secondary classification to filter out false positives.
The inputs are two successive grayscale image frames $I_{t}, I_{t-1}\in\mathbb{R}^{H\times W}$ where $H\times W$ are the dimensions of the input frames. 
We utilize the full image resolution of $2048 \times 2448$ during inference to maximize the chances of detection at long ranges. 
The outputs are a list of tracked objects with bounding box coordinates, track ID, 2D Kalman filtered state, estimated range, angular rate, and time to closest point approach (tCPA).
We then filter out false positive detections by projecting the ADS-B data into the image frame and ensuring that the projected ADS-B coordinates lie within a $100$ pixel radius of the predicted image bounding box.
Whenever we have false negatives from the model, we manually add in the bounding box label if the aircraft is visible in the image.

For the ADS-B data projection, we assume that the 4-camera setup is located at the origin of a north-east-down (NED) world reference frame where the positive z-axis points towards the earth's center and the positive x-axis points towards True North. Given all the relevant information, such as GPS coordinates and altitudes, we can compute the 3D position of aircraft using geodesics. The corresponding image frame location in homogeneous coordinates for camera $i$ ($i=1,\ldots,4$) is given by the following perspective projection:
$\mathbf{p}_i = \mathbf{K}_i[\mathbf{R}_i \mid \mathbf{t}_i] \mathbf{P}$, 
where $\mathbf{K}_i$ is a known $3\times 3$ intrinsic matrix for camera~$i$, $[\mathbf{R}_i|\mathbf{t}_i]$ constitutes the $3\times 4$ extrinsic matrix of camera $i$, and $\mathbf{P}\in\mathbb{R}^3$ is the aircraft 3D position in the world reference frame. In particular, we use the PnP-RANSAC algorithm to estimate $[\mathbf{R}_i|\mathbf{t}_i]$. Using time synchronization, the 2D correspondences for the recorded 3D observations are manually labeled using custom labeling software. Since the setup is static, this is a one-time calibration step that gives us the projection matrix for camera $i$, with which we can project new 3D aircraft data.

\subsection*{Trajectory and Weather Data Acquisition}

The trajectory and weather data are collected at both the KAGC and KBTP airports. Data collection follows similar procedures as our prior ADS-B dataset \cite{patrikar2021predicting}. We use a Stratux ADS-B receiver capable of receiving position reports on both the 1090 MHz and 978 MHz frequencies. The receiver was installed within the terminal area of each airport. The recording operation ran from 1:00 AM to 11:00 PM local time, a period selected to encapsulate the full range of aviation operations during both peak and off-peak hours. Data collection at KBTP started on September 18, 2020 and concluded on October 27, 2022. Data collection at KAGC began on October 31, 2021 and ended on February 17, 2023. Data was collected in discrete phases, resulting in 381 days or 36 million raw position reports of data at KBTP and 280 days or 27 million raw position reports of data at KAGC. Weather reports for the corresponding airports are collected post hoc in the form of METeorological Aerodrome Reports (METAR) strings. We use the Iowa State METAR repository~\cite{herzmann2004iowa} to compile reports for the duration of the ADS-B data. The total file size of the audio dataset is 1.9 GB compressed and 12 GB uncompressed.

\subsubsection*{Post-processing}   
The raw ADS-B data is first post-processed to remove corrupted and duplicate data points. The data is then filtered for altitude and distance from the airport. We nominally chose 6000 ft MSL and a 5 km radius around the airport. Once filtered, the data is transformed from a global to a local Cartesian coordinate frame in SI units with the origin at the end of the runway. The raw METAR strings are also processed to get wind velocity and direction relative to the runway in the local frame. Finally, we interpolate the trajectory data every second for all agents. TartanAviation provides the processed data along with the raw trajectory and weather data.

\subsection*{Speech Data}
The speech data is collected at both the KAGC and KBTP airports. The setup uses a Bearcat SR30C radio receiver capable of receiving aviation radio frequencies. For towered KAGC, the radio is tuned to receive 121.1 MHz, the air traffic control tower frequency for KAGC operations. KAGC has an active tower 24 hours a day. For KBTP, the radio was tuned to 123.05 MHz, the Common Traffic Advisory Frequency for KBTP operations. The audio is recorded at a rate of 44.1k samples per second onboard the system. The record trigger mechanism for the speech setup is similar in construction to the vision setup and uses ADS-B data to trigger recordings. The trigger threshold was set to 10 km. The corresponding raw ADS-B data is also provided for the communication recordings spanning multiple years for both airports. The communication data from KBTP starts on September 6 2020 and ends on October 27 2022. After filtering, this data comprises 278 days of data, corresponding to 790 GB of disk space. 
The communications data for KAGC starts on October 31, 2021 and ends on February 17, 2023. The KAGC data comprises 392 days of data, corresponding to 1358 GB of disk space. 
In total, the dataset contains 670 days of raw communications recordings. 
The total file size of the audio dataset is 2.15 TBs uncompressed and 505.2 GB compressed.

\subsubsection*{Post-processing}
We filtered the raw recordings using two conditions: audio clips must be longer than 1 second and have a maximum decibel level above -20 db. This removes erroneous recordings and clips that don't contain any spoken words.
The KAGC audio has 33289 audio files comprising 2131.9 hours of audio, with 327.714 hours above -20 db. The KBTP audio has 8534 files and 1242.9 hours of audio, with 149.9 hours above -20 db. This gives a total of 41823 files, 3374.8 hours of audio, and 477.6 hours of audio above -20 db.





\section*{Data Records}

The following sections present details on the data formats and provide information on file organization.

\begin{table}
\centering
\begin{tabular}{|l|l|l|}
\hline
\textbf{Extension} & \textbf{Nomenclature} & \textbf{Content}                     \\ \hline
.zip    &       \texttt{<camera\_id>\_<timestamp>}    & Zip folder containing sequence data \\ \hline
$\hookrightarrow$ .mp4     &   \texttt{<camera\_id>\_<timestamp>}    &     Video File   \\ \hline
$\hookrightarrow$.avi   &    \texttt{<camera\_id>\_<timestamp>\_sink\_verified}      &  Video File with embedded labels   \\ \hline
$\hookrightarrow$.srt   & \texttt{<camera\_id>\_<timestamp>\_subtitle}  &       Raw timestamps of the recorded video                               \\ \hline
$\hookrightarrow$.pkl   & \texttt{<camera\_id>\_<timestamp>\_sink\_adsb}   &  Raw ADS-B dictionary  \\ \hline
$\hookrightarrow$.pkl   & \texttt{<camera\_id>\_<timestamp>\_acft\_sink}   &  Raw aircraft type data  \\ \hline
$\hookrightarrow$.zip    &     \texttt{<camera\_id>\_<timestamp>\_labels}   &  Zip folder containing the image labels         \\ \hline
$\hookrightarrow$$\hookrightarrow$.label    &     \texttt{<frame\_number>.label}   &  Text file containing label data  \\ \hline

\end{tabular}
\caption{File structure information for TartanAviation image data}
\label{tab:vis_data}
\end{table}

\subsection*{Image Data}

The image dataset is split across $550$ independent sequences. We define a sequence as all of the data recorded during a single event where the camera recordings were started and stopped. The vision data folder contains multiple zipped files, each associated with a particular camera recording for that sequence. Each zipped sequence folder has multiple files, as presented in Table~\ref{tab:vis_data}. Further information regarding the nomenclature and file contents is also shown in Table \ref{tab:vis_data}. In addition to the video files and labels, we also provide ADS-B data for each sequence.

\subsection*{Trajectory and Weather Data}

TartanAviation provides both raw and processed data for each airport. Raw data is separated into individual folders for each day of collection. Each raw data folder has CSVs with fields detailed in Table \ref{tab:adsb}. The processed files are available as comma-separated TXT files with fields described in \ref{tab:adsb}.

\subsection*{Speech Data}
 



Both the raw and filtered audio files are included in the dataset. The filtered data is organized in a directory structure by location, year, month, and day. Each day is individually zipped, contains audio files in the WAV format, and has an accompanying text file that contains the audio clip's start, end, and total time.





\begin{table}[]
\begin{tabular}{|lll|lll}
\hline
\multicolumn{3}{|c|}{\textbf{Raw Data Fields}}                                                                        & \multicolumn{3}{c|}{\textbf{Processed Data Fields}}                                                                                                                                                                   \\ \hline
\multicolumn{1}{|l|}{\textbf{Field}} & \multicolumn{1}{l|}{\textbf{Units}}  & \textbf{Description}                    & \multicolumn{1}{l|}{\textbf{Field}}        & \multicolumn{1}{l|}{\textbf{Units}}       & \multicolumn{1}{l|}{\textbf{Description}}                                                                                    \\ \hline
\multicolumn{1}{|l|}{ID}             & \multicolumn{1}{l|}{\#}              & ADS-B Aircraft ID                       & \multicolumn{1}{l|}{Frame}                 & \multicolumn{1}{l|}{\#}                   & \multicolumn{1}{l|}{Relative Timestep}                                                                                       \\ \hline
\multicolumn{1}{|l|}{Time}           & \multicolumn{1}{l|}{HH:MM:SS.ss}     & Time of observation                     & \multicolumn{1}{l|}{ID}                    & \multicolumn{1}{l|}{\#}                   & \multicolumn{1}{l|}{ADS-B Aircraft ID}                                                                                       \\ \hline
\multicolumn{1}{|l|}{Date}           & \multicolumn{1}{l|}{MM/DD/YYY}       & Date of observation                     & \multicolumn{1}{l|}{x}                     & \multicolumn{1}{l|}{km}                   & \multicolumn{1}{l|}{Local X Cartesian Position}                                                                              \\ \hline
\multicolumn{1}{|l|}{Altitude}       & \multicolumn{1}{l|}{Feet}            & Aircraft Altitude (Mean Sea Level)      & \multicolumn{1}{l|}{y}                     & \multicolumn{1}{l|}{km}                   & \multicolumn{1}{l|}{Local Y Cartesian Position}                                                                              \\ \hline
\multicolumn{1}{|l|}{Speed}          & \multicolumn{1}{l|}{Knots}           & Aircraft Speed                          & \multicolumn{1}{l|}{z}                     & \multicolumn{1}{l|}{km}                   & \multicolumn{1}{l|}{Local Z Cartesian Position}                                                                              \\ \hline
\multicolumn{1}{|l|}{Heading}        & \multicolumn{1}{l|}{Degrees}         & Aircraft Heading                        & \multicolumn{1}{l|}{\multirow{2}{*}{$w_x$}} & \multicolumn{1}{l|}{\multirow{2}{*}{m/s}} & \multicolumn{1}{l|}{\multirow{2}{*}{\begin{tabular}[c]{@{}l@{}}Component of wind along\\  the dominant runway\end{tabular}}} \\ \cline{1-3}
\multicolumn{1}{|l|}{Lat}            & \multicolumn{1}{l|}{Decimal Degrees} & Latitude of the aircraft               & \multicolumn{1}{l|}{}                      & \multicolumn{1}{l|}{}                     & \multicolumn{1}{l|}{}                                                                                                        \\ \hline
\multicolumn{1}{|l|}{Lon}            & \multicolumn{1}{l|}{Decimal Degrees} & Longitude of the aircraft              & \multicolumn{1}{l|}{$w_y$}                  & \multicolumn{1}{l|}{m/s}                  & \multicolumn{1}{l|}{\begin{tabular}[c]{@{}l@{}}Component of wind across\\  the dominant runway\end{tabular}}                 \\ \hline
\multicolumn{1}{|l|}{Age}            & \multicolumn{1}{l|}{Seconds}         & Time since last observation             &                                            &                                           &                                                                                                                              \\ \cline{1-3}
\multicolumn{1}{|l|}{Range}          & \multicolumn{1}{l|}{km}              & Distance from airport centre            &                                            &                                           &                                                                                                                              \\ \cline{1-3}
\multicolumn{1}{|l|}{Bearing}        & \multicolumn{1}{l|}{Degrees}         & Bearing Angle with respect to North     &                                            &                                           &                                                                                                                              \\ \cline{1-3}
\multicolumn{1}{|l|}{Tail}           & \multicolumn{1}{l|}{}                & Aircraft Registration Number            &                                            &                                           &                                                                                                                              \\ \cline{1-3}
\multicolumn{1}{|l|}{AltisGNSS}      & \multicolumn{1}{l|}{boolean}         & Indicator flag for Altitude Measurement &                                            &                                           &                                                                                                                              \\ \cline{1-3}
\end{tabular}
\caption{Variable description for the TartanAviation ADS-B trajectory data.}
\label{tab:adsb}
\end{table}



\section*{Technical Validation}
\begin{figure}
    \centering
    \includegraphics[width=0.85\linewidth]{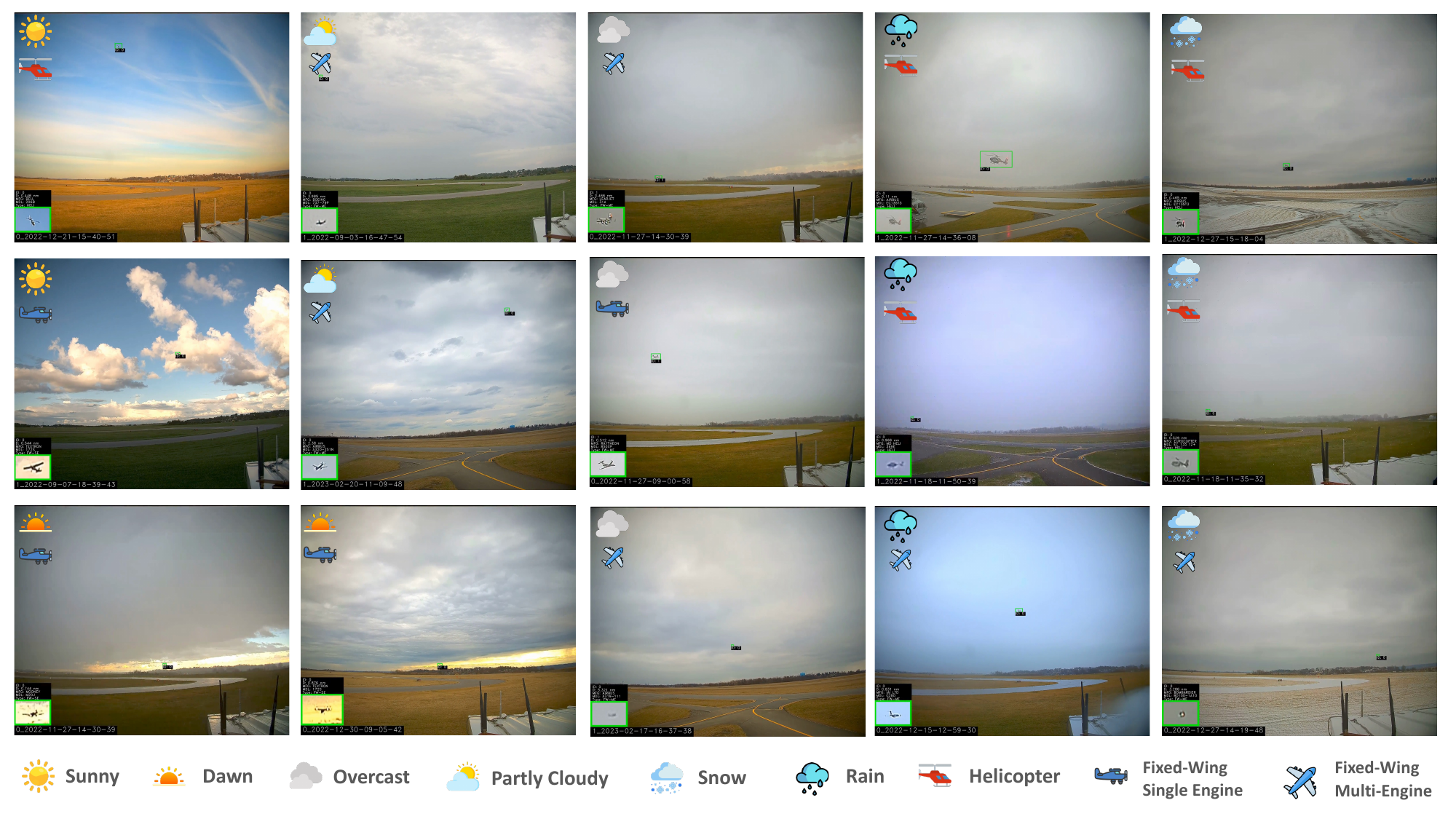}
    \caption{Qualitative samples from the TartanAviation Image dataset showcasing the diversity of the collected images in different lighting conditions, seasons, cloud covers, and aircraft types.}
    \label{fig:kagc_vision_samples}
\end{figure}
\begin{figure}
    \centering
    \begin{subfigure}{0.45\textwidth}
        \centering
        \includegraphics[width=0.75\linewidth]{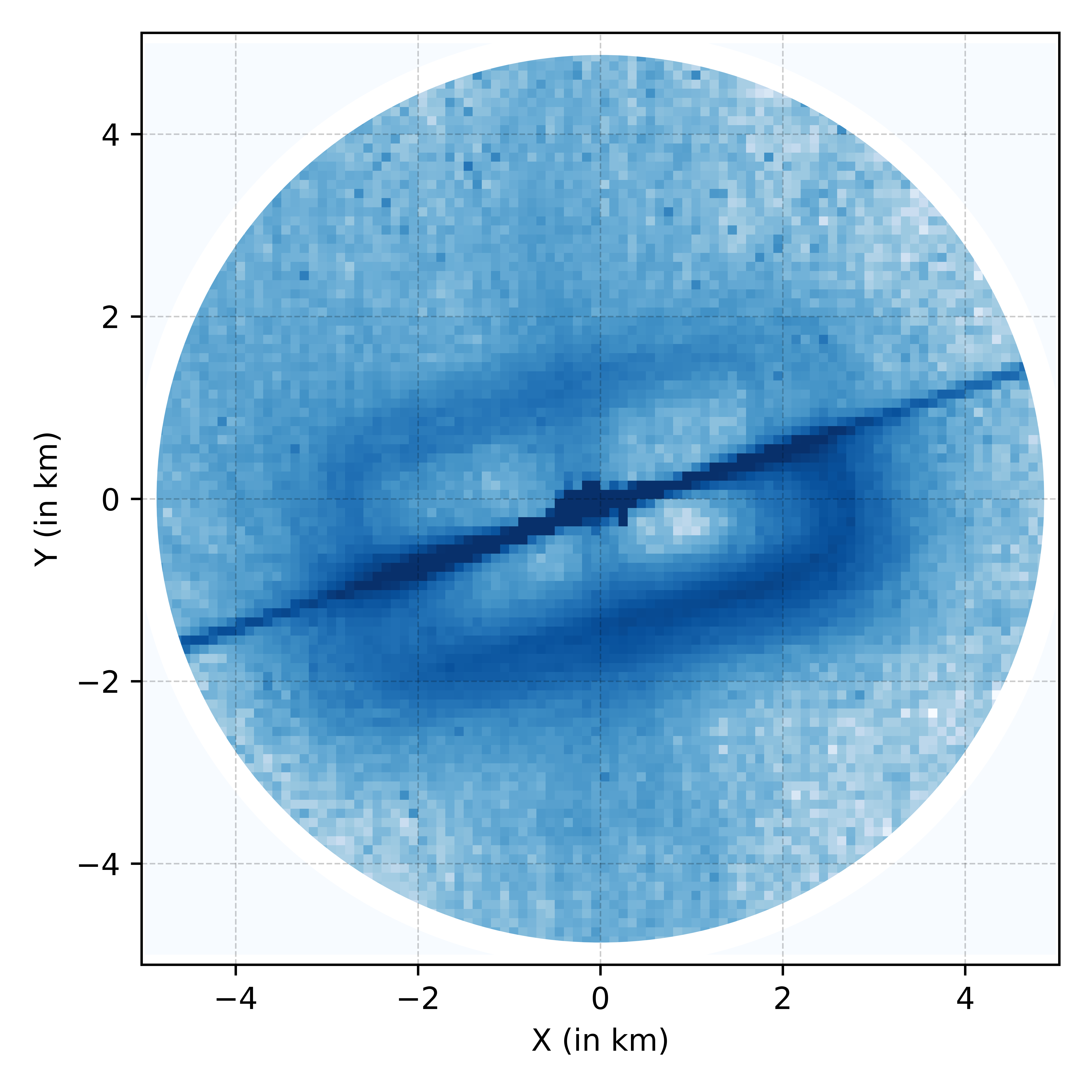}
        \caption{Pittsburgh-Butler Regional Airport}
        \label{fig:kbtp_traj}
    \end{subfigure}
    \hfill
    \begin{subfigure}{0.45\textwidth}
        \centering
        \includegraphics[width=0.75\linewidth]{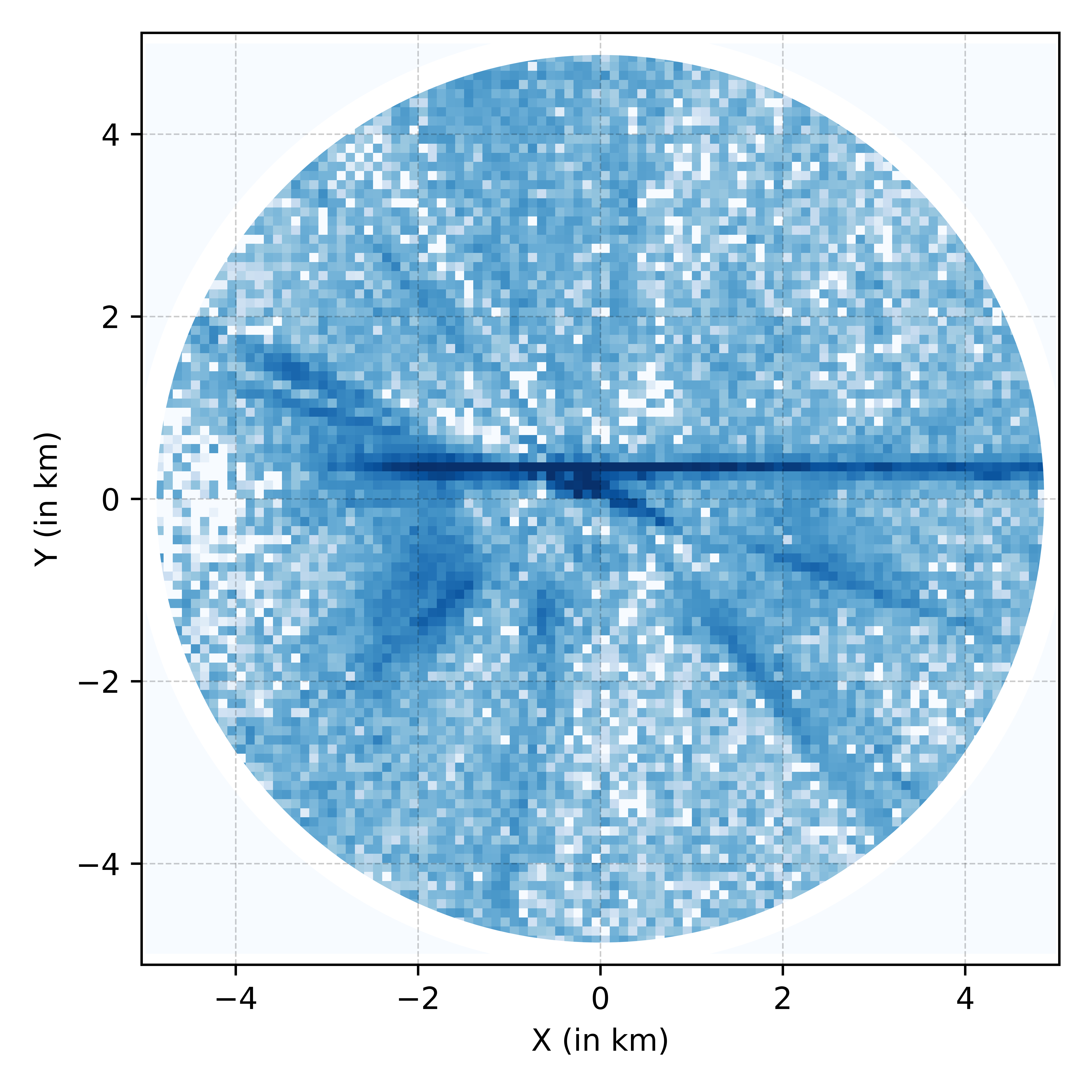}
        \caption{Allegheny County Airport}
        \label{fig:kagc_traj}
    \end{subfigure}
    \caption{Log-normed trajectory histograms from ADS-B aircraft position reports.}
    \label{fig:trajs_kagc_kbtp}
\end{figure}
\begin{figure}
    \centering
	\includegraphics[width=0.9\columnwidth]{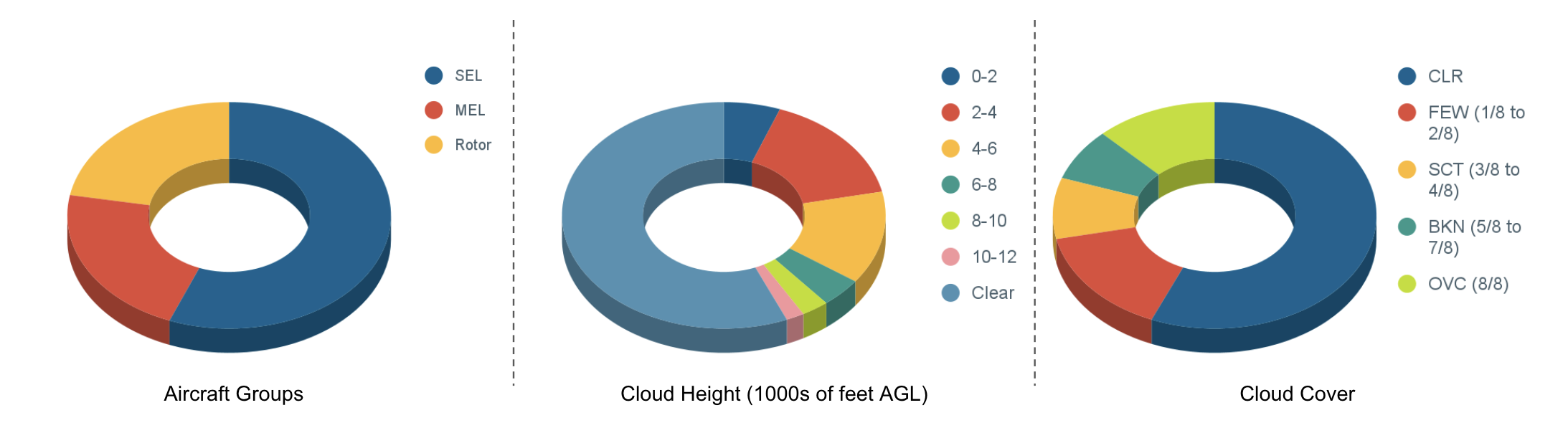}
	\caption{Quantitative diversity from the TartanAviation Image dataset showcasing the distribution of the collected images with respect to aircraft groups, cloud heights, and cloud cover.}
 \label{fig:kagc_analysis}
\end{figure}
The data collected were assessed to ensure the reliability of the data provided for each modality individually.

\subsection*{Image Data}
\begin{figure}[h!]
    \centering
    \begin{subfigure}{0.90\textwidth}
        \centering
        \includegraphics[trim={0.3cm 0cm -2.9cm 0cm},clip,width=\columnwidth]{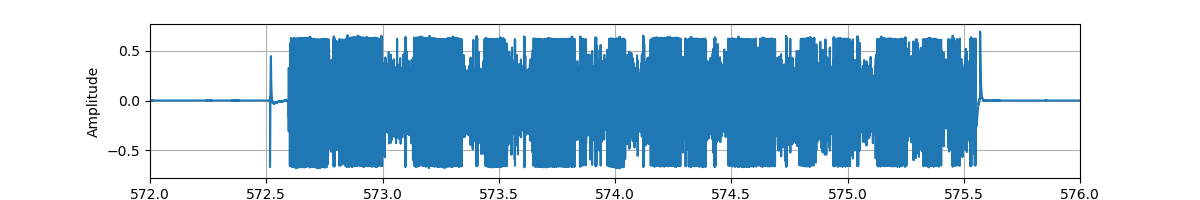}
    \end{subfigure}
    \begin{subfigure}{0.90\textwidth}
        \centering
        \includegraphics[trim={1cm 0cm 3cm 1.1cm},clip,width=\columnwidth]{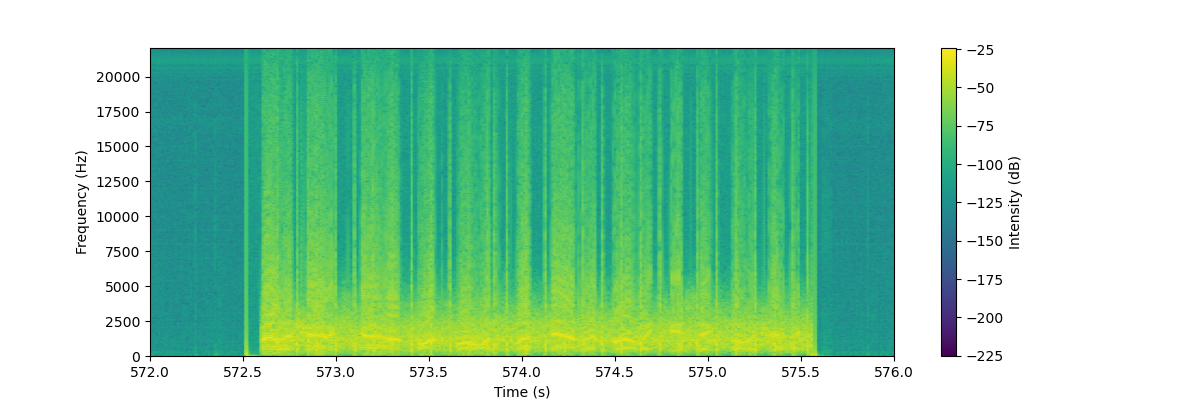}
    \end{subfigure}
    \caption{A portion of the spectrogram and waveform for audio file 6.wav from KBTP on November 2, 2020. The spectrogram shows the frequency of the audio signal versus time. In the time period of this figure, it is clear when the pilot is speaking, which can be seen as the higher intensity portion of the spectrogram and the spikes in amplitude in the waveform.}
    \label{fig:spectrogram}
\end{figure}
Figure \ref{fig:kagc_vision_samples} shows example images from the dataset. The subset highlights the variation in lighting conditions, seasons, cloud covers, cloud heights, and aircraft types. The tightly labeled bounding boxes provided with each image are also shown. The scale of the bounding boxes with respect to the total image size highlights the challenge faced by object detection algorithms trying to reliably detect aircraft in a cluttered background. Figure \ref{fig:kagc_analysis} shows quantitative results for the entire dataset. To showcase the diversity in aircraft, we group the images based on single-engine land (SEL), multi-engine land (MEL), and Rotorcraft (rotor). This shows an almost equal distribution in the dataset with SEL as the major class with 56\%. Further, we group the images based on the recorded cloud height. Cloud height has a direct impact on the available lighting. While 56.4\% of the images have no clouds in them, 5.6\% of images have very low cloud layers below 2000 feet above ground level (AGL). Grouping the images by cloud coverage, we observe that while 56.4\% of data has no clouds, 12.2\% of data has overcast skies. In addition to this, 6.4\% of data has active precipitation while 12.4 \% of data has visibility less than 10 statute miles.


\subsection*{Trajectory Data}

Figure~\ref{fig:trajs_kagc_kbtp} showcases trajectory histograms for the KAGC and KBTP Airports. The histograms represent the aircraft occupancy frequencies on a log scale around both airports. The airport is at the center in both images, with the geographic north pointing upwards. The effect of runway geometries is clearly visible. Figure \ref{fig:kbtp_traj} shows the $08/26$ KBTP runway while Figure \ref{fig:kagc_traj} shows the crossing smaller $13/31$ and larger $10/28$ runways at KBTP. Also clearly visible are the different types of operations at both airports. KBTP is an un-towered airport and home to a few flight schools. The left traffic patterns for both runways are clearly visible, highlighting the adherence to FAA guidelines when operating in an un-towered airfield. KAGC, on the other hand, is a towered airfield that hosts medical evacuation helicopters and business jet traffic in addition to flight schools. This leads to more straight-line arrivals and departures from the airport runways, as reflected in the figure.      


\subsection*{Speech Data}

Our filtering of the audio data ensures that empty files are discarded while any that could contain speech data are included. The threshold of -20 db was chosen to preserve any audio louder than radio silence. Figure \ref{fig:spectrogram} shows a portion of the spectrogram and waveform of one of the audio files from our dataset, specifically 6.wav from KBTP on November 2, 2020. It is apparent in the spectrogram where the pilot is speaking and where there is radio silence. Furthermore, some low-intensity noise during radio silence occurs at clear intervals and shows up as vertical lines with even spacing. During the speaking portion of the audio, there is a clear spike in intensity between all bands, with the highest being between 0-3 kHz. The human ear is highly sensitive in frequencies between 1-4 kHz, with the sensitivity dropping off steeply to the limit of around 20 kHz \cite{Sivian_White_1933}. Research in the interpretability of speech finds that the values below 4 kHz yield high accuracy in articulation \cite{French_Steinberg_1947}. The spectrogram shows these vital frequencies for speech data are recorded sufficiently. 



\section*{Usage Notes}



In generating and post-processing TartanAviation, care was taken to organize data in formats commonly accepted by respective communities. This leads to hassle-free integration with existing dataloaders for specific tasks like vision-based small object detection and trajectory prediction. Pytorch Dataloaders for vision-based object detection are also provided in the supplementary codebase released with the manuscript. The data is available on high throughput servers without a paywall or any other access restrictions to enable widespread utilization. The provided download scripts ensure that the dataset can be downloaded in user-specified chunks to enable processing on local machines as well as large-scale servers. The modalities can be used independently or in conjunction. TartanAviation not only challenges the existing methods for established tasks like speech-to-intent prediction, object detection, and trajectory prediction but also opens up exciting avenues for leveraging multi-modal data in the context of aviation.

\section*{Code availability}
The raw and processed data for all the modalities is available at \href{https://theairlab.org/tartanaviation}{https://theairlab.org/tartanaviation/}. The scripts to record, post-process and download each modality are publicly available at \href{https://github.com/castacks/TartanAviation.git}{https://github.com/castacks/TartanAviation.git}.

\bibliography{sample}


\section*{Acknowledgements}

This work is supported by the Mitsubishi Heavy Industries (MHI) project \#A025279. This material is based upon work supported by the National Science Foundation Graduate Research Fellowship under Grant No. DGE1745016. We would like to thank the Allegheny County Airport Authority and the Office of Manager of Pittsburgh-Butler Regional Airport for their support.

\section*{Author contributions}
The data collection was conceived by S.S. The hardware setup was created by J.P. The audio and trajectory recording scripts were created by J.P. The image recording scripts were created by S.G. The image data was labeled and post-processed by M.H. The trajectory data was post-processed and analyzed by J.D. The audio data was post-processed and analyzed by B.M. All authors assisted in setup maintenance and data retrieval. All authors reviewed the manuscript.


\section*{Competing interests}

The authors declare no competing interests.

\end{document}